# Benchmarking Large Language Models for Handwritten Text Recognition

Giorgia Crosilla[1], Lukas Klic[2], and Giovanni Colavizza[3,4]

[1]University of Bologna, Bologna, Italy

giorgia.crosilla@studio.unibo.it

[2]I Tatti, The Harvard University Center for Italian Renaissance Studies, Florence, Italy

lklic@itatti.harvard.edu

[3]University of Bologna, Department of Philology and Italian Studies, Bologna, Italy.

[4]University of Copenhagen, Department of Communication, Copenhagen, Denmark

giovanni.colavizza@unibo.it; colavizza@hum.ku.dk

**Abstract**

**Purpose** - The aim of this work is to provide an overview of the current capabilities of Multimodal Large Language Models (MLLMs) for Handwritten Text Recognition (HTR), assessing their potential when compared to traditional task-specific, supervised models.

**Design/Methodology/Approach** - The approach is that of using a set of openly available benchmarks to compare different LLMs with strong task-specific supervised baselines for the task of HTR.

**Findings** - The results show that LLMs currently show a strong performance on English texts, yet they demonstrate a weaker performance on languages other than English, and do not possess a significant capability for self-correction. Moreover, their comparison with *Transkribus*'s models highlight that proprietary LLM models are the best performing, in particular on modern handwriting, while for historical documents the overall performance comparison between LLMs and *Transkribus* is not consistent.

**Originality/Value** - The authors are not aware of a similar study relying on open benchmarks.

## 1 Introduction

In recent years, advancements in Generative AI models have begun to influence research in the Digital Humanities. The capabilities of Large Language Models (LLMs) manifest both in their ease of use for scholars, and in the wide range of applications that Generative AI can support within knowledge organization, such as the ability to classify, process, and extract meaningful information from cultural data, support learning and teaching activities, facilitate collaborative annotation and enrichment of texts, and assist in interpretative and critical analysis. All these aspects, which ultimately enhance data accessibility, are facilitated by the models' capacity to handle and analyse vast amounts of information. As a result, it becomes possible to carry out large-scale analyses in an unprecedented manner within the humanities, offering broader and faster insights thanks to the models' computational speed. In this sense, the increasing interest in the application of Generative AI within the Digital Humanities is not directed toward replacing human-driven research, but rather toward serving as a support tool that can enhance scholarly inquiry.

Within this broader context, the paper aims to explore the applicability of Large Language Models (LLMs) in the field of Handwritten Text Recognition (HTR). Automated Text Recognition (ATR), which comprehends both HTR and Optical Character Recognition (OCR), has been proposed as a solution to manual transcriptions which require significant human effort and extensive time to process large quantities of data. However, unlike OCR, which is characterized by standardized printed fonts, HTR possesses numerous challenges from the variability of handwriting styles, historical calligraphic conventions, and various page deterioration issues, such as ink marks, bleed-through, and stains. For this reason, despite its potential, HTR has struggled to meet the same quality standards as OCR on printed texts. As a result, handwritten transcriptions are frequently overlooked during digitization processes (Ströbel, 2023, p.46) and, consequently, they are not as often integrated in digital libraries (Terras, 2022, p.181). The progressive improvement in HTR performance could enhance organization, accessibility, and information retrieval within digitized collections, focusing on accessibility based on the documents' contents, rather than archival descriptions (Colavizza *et al.*, 2022, pp.7-8). Moreover, integrating HTR into collections could shift the perception of cultural objects from being mere digital entities to valuable data, offering a starting point for further analysis (Nockels *et al.*, 2024, p.150), visualizations, and research opportunities based on handwritten material (Muehlberger *et al.*, 2019, p.956).

So far, machine learning models applied to HTR have relied on a supervised learning workflow requiring a significant amount of labelled data to train specialized models for peculiar handwriting styles. This dependency on large, high-quality annotated datasets poses a challenge, particularly for historical materials where such resources are often scarce. Additionally, traditional HTR methods struggle with adaptability, as models trained on specific scripts, languages, or time periods frequently require retraining when applied to new data. The workflow itself is both time-consuming and error-prone due to the separation between layout analysis and text recognition, where mistakes in one stage can negatively impact the other. Moreover, many traditional approaches operate at different granularity levels without a deep understanding of semantic context, limiting their effectiveness in handling ambiguous handwriting or degraded documents. These limitations are also evident in widely used HTR platforms that rely on such models including *Transkribus*[1], one of the most accessible and popular for this task, even among non-experts.

Considering these aspects and recent developments in deep learning related to Multimodal Large Language Models (MLLMs), this research primarily assesses their suitability for fulfilling the main goal of HTR: the application of general models to successfully recognize the content of modern and historical multilingual handwritten materials (Ströbel, 2023, p.13), while diminishing costs and manual supervision. Specifically, the study aims at answering four research questions, providing insight into the current state of LLMs applied to HTR, with comparisons to *Transkribus*:

- What is the level of accuracy of MLLMs in transcribing multilingual handwritten text?
- Do proprietary models and small open-source models perform similarly?
- Can MLLMs autocorrect and improve their previous predictions?
- How do the results produced by MLLMs compare to those of the models available on *Transkribus*?

The first research question aims to evaluate if adapting MLLMs to HTR, using a workflow based on users' interaction with natural language, surpasses the traditional approach focused on developing *ad hoc* models. The comparison between proprietary and open models aims to determine if a viable free small-scale open-source alternative can achieve comparable accuracy and reduce costs. The capability of post-correction is introduced both to assess if the model is capable of "reasoning" on the previous results and to further reduce the manual effort needed. Furthermore, the comparison with *Transkribus* provides insight into how general models perform when compared to specialized ones published by one of the widest used HTR user- friendly platforms. This article is structured as follows: first, a brief overview of the literature and related applications of LLMs to HTR is presented. The methodology is then outlined, detailing the workflow, datasets, models, setup, prompt engineering, and evaluation metrics. Finally, the results of both the zero-shot and self-correction tasks are discussed.

# 2 Related Works

## *2.1 Supervised Learning*

HTR has traditionally relied on two main supervised approaches: training models from scratch or fine-tuning pre-existing architectures. However, these approaches required the use of two separate models: one for layout analysis and line segmentation, and another for the actual text recognition (AlKendi *et al.*, 2024, pp.4-5). This creates an error-prone pipeline in which mistakes made during the layout analysis step may go undetected and, if not manually corrected, are propagated through the subsequent recognition stages. Moreover, manual effort is required in image pre-processing (i.e. binarization and skewing), along with the creation of ground truth layout and textual annotations. The annotation granularity may vary depending on whether recognition is performed at the word, line, paragraph, or page level.

[1] https://www.transkribus.org/ (accessed 29/01/2025).

In the case of historical documents, line-level recognition has been especially favoured, due to complex layouts and limited training data. Early approaches relied on stochastic methods (Wuthrich *et al.*, 2009), followed by Multidimensional Long Short-Term Memory-Recurrent Neural Networks (MLSTM-RNN) with Connectionist Temporal Classification (CTC) (Puigcerver, 2017; Voigtlaender *et al.*, 2016), and more recently by Transformer-based models (Kang *et al.*, 2020; Li *et al.*, 2023, 2024; Ströbel *et al.*, 2022; Wick *et al.*, 2021). In this context, various HTR toolkits have emerged, such as *PyLaia*[2] and *Kraken*[3], which serve as the foundational frameworks for the models employed by the user-friendly platforms *Transkribus* and *eScriptorium*[4], respectively.

A major shift occurred around 2016 with the introduction of attention mechanisms, which enabled the development of models capable of recognizing entire paragraphs (Bluche *et al.*, 2016) and full-pages (Carbonell *et al.*, 2019; Chammas *et al.*, 2018; Coquenet *et al.*, 2021; Moysset *et al.*, 2017; Tensmeyer and Wigington, 2019; Wigington *et al.*, 2018; Yousef and Bishop, 2020). Further advancements have been achieved by implementing Transformer-based architectures (Coquenet *et al.*, 2023; Singh and Karayev, 2021). These approaches aimed to more closely replicate real-world scenarios, where full documents are processed without prior segmentation. The transition to end-to-end architectures marked a significant step toward reducing the dependency on large quantities of labelled data while increasing the flexibility and scalability of HTR systems.

## *2.2 New approaches in HTR using LLMs*

Even though the HTR state-of-the-art models are pre-trained on a wide number of handwritten texts only, the ease of interaction with general purpose MLLMs raises the question of their applicability in the HTR field. By merging layout analysis and text recognition, MLLMs can be adapted to the task of HTR by providing a well-defined prompt and the raw image, contributing to a simpler and faster workflow. Textual and contextual understanding are not limited to the analysis of the text but through “cross-modal semantic understanding” where tables and illustrations play a key role for deciphering the content of a document (Liu *et al.*, 2024, p.4). The most straightforward approach to apply MLLMs to HTR is by using a zero-shot approach, as demonstrated in the following related works.
First, L. Li (2024) applied MLLMs in HTR task by comparing gemini-pro-vision[5] model with “traditional models” based on CNN-BiLSTLM and *TrOCR* (Li *et al.*, 2023). The comparison has been undertaken using publicly available datasets such as ICDAR (2014-2017) for English, French, and German. While traditional methods achieved the best CER results, Gemini exhibited a performance gap between English and other languages, indicating a language bias.

Moreover, Kim *et al.*, (2025) focused on the comparison between Claude Sonnet 3.5 and GPT-4o against traditional OCR and HTR systems such as *EasyOCR*[6], *Keras*[7], *PyTesseract*[8], and *TrOCR* to process handwritten French tabular data. Using progressively complex prompts, they assessed performance at both the line and full-page levels. For full-document recognition, Claude Sonnet 3.5 achieved the highest accuracy, surpassing traditional methods.

The paper closer to this research is the one by Humphries *et al.*, (2024) which provides a comparison between different proprietary models Claude Sonnet 3.5, Gemini-1.5 Pro-002 and gpt-4o-06-08- 2024, along with a *Transkribus PyLaia*[9] model and “The Text Titan I” supermodel. The paper investigates the accuracy of LLM in HTR using a zero-shot approach, a self-correction approach, a correction using a different model and an LLM correction of *Transkribus* outputs. While the study presents a valid methodology and comparisons for assessing the applicability of LLMs in HTR research, two key limitations must be noted. First, it considers only English texts, which may introduce a shortcoming given that most LLM training data is in English. Consequently, the results cannot be regarded as representative of the models’ recognition capabilities, as no datasets in languages other than English were analyzed. Second, the study does not use public datasets, arguing that they fail to “simulate real-world conditions” and that LLMs may have been exposed to available HTR data during training, potentially leading to artificially low error rates (Humphries *et al.*, 2024, p.12). Although this concern is valid in principle, the study does not provide details on the prompts used or the results obtained. Furthermore, the test was conducted only on Gemini, with no additional evaluations on other models to substantiate this claim.

[2] https://github.com/jpuigcerver/PyLaia (accessed 05/01/2025).
[3] https://github.com/mittagessen/kraken/tree/main (accessed 05/01/2025).
[4] https://escriptorium.inria.fr/ (accessed 05/01/2025).
[5] https://deepmind.google/technologies/gemini/pro/ (accessed 05/01/2025).
[6] https://github.com/JaidedAI/EasyOCR (accessed 05/02/2025).
[7] https://keras-ocr.readthedocs.io/en/latest/ (accessed 05/02/2025).
[8] https://github.com/h/pytesseract (accessed 05/02/2025).
[9] https://help.Transkribus.org/choosing-a model?hstc=48127928.e0294bbb9dc7803f11d898931d9f9177.1720904643552.1734642343038.1735152933219.28& hssc=48127928.1415.1735152933219& hsfp=1436914090 (accessed 29/01/2025).

As for post-correction strategies, previous studies have demonstrated that LLMs are unable to correct their own prediction, sometimes providing worst answers than the initial ones (Stechly *et al.*, 2023). This can be caused by the fact that LLMs are built to output the most probable token in a sequence, and "cannot properly judge the correctness of their reasoning" (Huang *et al.*, 2024, p.4). While in the OCR domain recent research has reported notable improvements through self-correction mechanisms (Bourne, 2025; Greif *et al.*, 2025) similar attempts in the HTR field have not yielded comparable results (Humphries *et al.*, 2024).

# 3 Methodology

## *3.1 Benchmark Workflow*

The main difference between the previous approaches and MLLMs lies in the user's interaction and in the simplification of the workflow. In the supervised environment, human involvement is required not only in image pre-processing and annotation but also in the post-correction of layout detection before text recognition. MLLMs substantially reduce this complexity by accepting unannotated images, processing layout, and text recognition at the same time. As a consequence, the creation of Ground Truth (GT) may not be needed for anything other than computing the accuracy of the model. Moreover, instructions are articulated using natural language, which facilitates the immediate interaction and task refinement from a user's perspective. Overall, MLLM's reduce manual supervision and processing time, while proposing a more approachable setting to non- expert users.

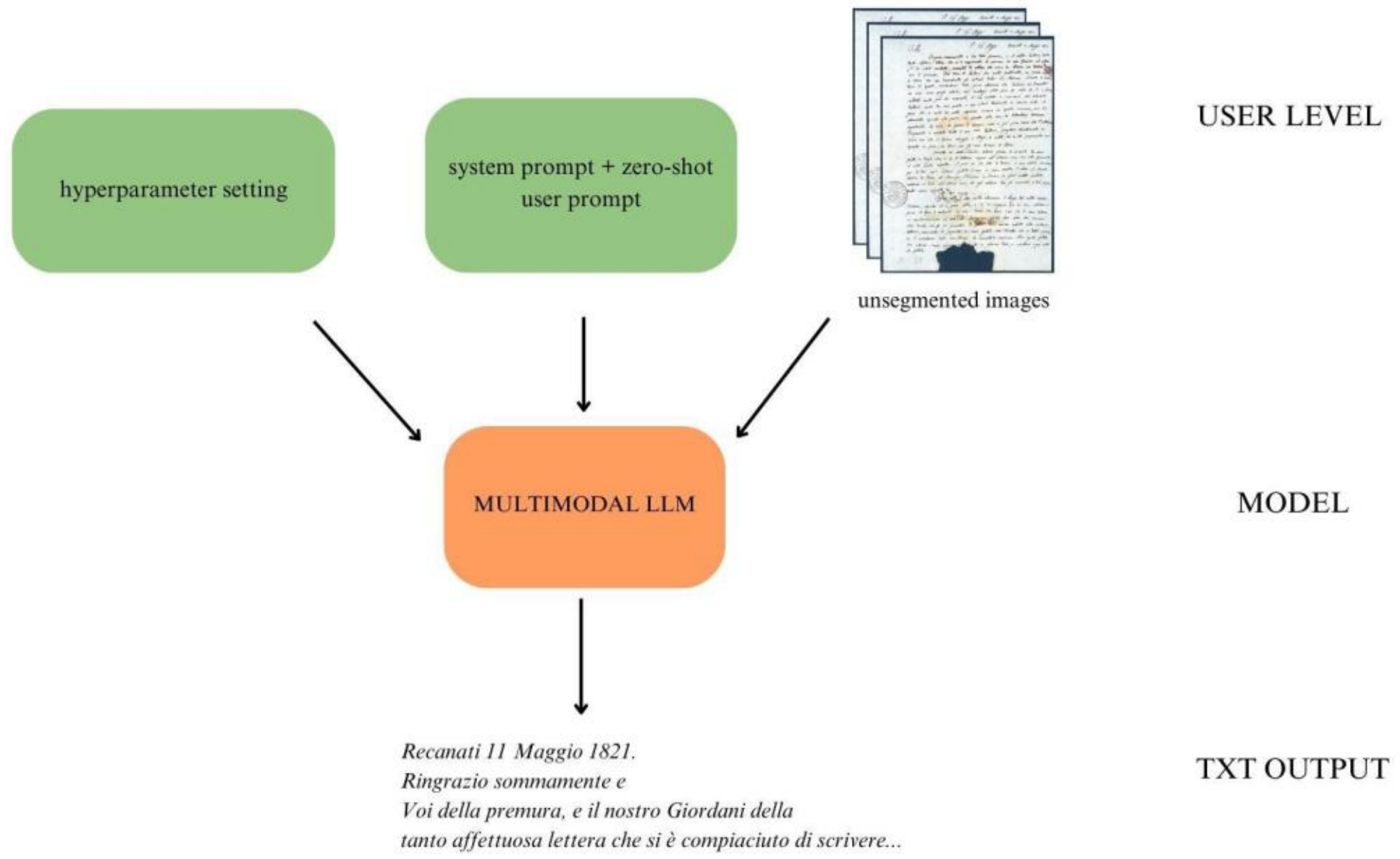

Figure 1: Application of MLLMs in a zero-shot approach to Handwritten Text Recognition from a user perspective.

The application of LLMs for the HTR task involves a different workflow than the one based on pre-processing, segmentation, and text recognition. While preprocessing may be required in some cases where models require image resizing or corrections to orientation, segmentation is no longer necessary, nor are specific GT annotations related to both text and layout. In this research, to ensure a fair comparison, the same datasets, implementation strategies, hyperparameters, and prompts are used across all models. Models are evaluated using the *Transkribus* platform, API access for proprietary models, or local installation for open-source models. The study also investigates the potential of incorporating LLMs as a post-correction step in the workflow, to assess whether LLMs could detect and correct errors in their initial predictions and potentially improve accuracy. Finally, models' predictions are evaluated using traditional metrics commonly found in literature, such as Character Error Rate (CER), Word Error Rate (WER), and Bag-of-Words Word Error Rate (BoW-WER), guaranteeing a comparable evaluation with previous studies that tested models' performance on the same datasets.

## 3.2 Models

The benchmarking project was done using eight MLLMs: three proprietary models (GPT-4o[10], GPT-4o-mini[11]and Claude 3.5 Sonnet[12]) and five different small open-source models (MiniCPMV- 2 6[13], Qwen2-vl- 7B[14], Pixtral12B[15], InternVL2-8B[16] and Phi-3-mini-instruct-128k[17]). They are characterized by multilingual support, along with implementation via API or local installation. They represent state-of-the-art multilingual proprietary and small open-source multimodal MLLMs as of October 2024.

The models already available on *Transkribus* are based on *PyLaia* and TrHTR[18]. The former involves a supervised approach to create a model from scratch or to fine-tune an existing model. Instead, "supermodels", are built on the TrHTR Transformer architecture which leverages the one of *TrOCR*. These models, available only via paid subscription, have been pre-trained on extensive datasets and are not intended for fine-tuning (Terras *et al.*, 2025, p.21).
In this application, the supermodel "The Text Titan I", trained on 16th to 21st century multilingual material was chosen for the detection of English, German and French, reaching an average CER of 2,95%. On the other hand, since no supermodels are currently available for Italian, the *PyLaia*-based "Italian Handwriting M1" was used. This is a model released by the *Transkribus* team which recognizes handwriting from the 16th to the 19th century and achieves an average CER of 6.70%.

## 3.3 Datasets

The datasets selection has been undertaken considering their availability on the web, open-source status, the full-page accessibility, the documentation provided in a related paper and the usage in literature[19]. The criteria were based on a table published by Cascianelli *et al.*, (2022, p.2), where the highly used benchmark datasets in literature are reported. However, in that table most of the mentioned datasets are suited only for line-level recognition. Considering this as the primary source, the research for other page- level datasets was further expanded in literature, considering, when indicated, the already provided splitting to ensure our approach was comparable to previous methods (see Table I.).

| **Dataset** | **Language** | **Period** | **Validation set (pages)** | **Validation Set** | **Available Dataset (URL)** | **Paper** |
|---|---|---|---|---|---|---|
| **IAM** | English | Modern | 116 (Aachen Split) | Provided | IAM dataset (accessed 14/01/2025) | (Marti and Bunke, 1999, 2002) |
| **RIMES** | French | Modern | 100 (DVD1 split) | Last 100 images of training set | RIMES dataset (accessed 14/01/2025) | (Augustin *et al.*, 2006) |
| **LAM** | Italian | 18th Century | 171 | 10% of the Training set | LAM dataset (accessed 14/01/2025) | (Cascianelli *et al.*, 2022) |
| **Leopardi** | Italian | 19th Century | 16 | Provided | Leopardi dataset (accessed 14/01/2025) | (Cascianelli *et al.*, 2021) |
| **Bentham** | English | 18th–19th Century | 50 | Provided | Bentham dataset (accessed 14/01/2025) | (Sanchez *et al.*, 2016) |
| **READ2016** | German | 15th–19th | 50 | Provided | READ2016 | (Sanchez *et al.*, 2016) |

[10] https://openai.com/index/gpt-4/(accessed 29/01/2025).
[11] https://openai.com/index/gpt-4o-mini-advancing-cost-efficient-intelligence/ (accessed 29/01/2025).
[12] https://www.anthropic.com/news/claude-3-5-sonnet (accessed 29/01/2025).
[13] https://huggingface.co/openbmb/MiniCPM-V-2_6 (accessed 29/01/2025).
[14] https://huggingface.co/Qwen/Qwen2-VL-7B-Instruct (accessed 29/01/2025).
[15] https://mistral.ai/en/news/pixtral-12b(accessed 29/01/2025).
[16] https://huggingface.co/OpenGVLab/InternVL2-8B (accessed 29/01/2025).
[17] https://huggingface.co/microsoft/Phi-3-mini-128k-instruct(accessed 29/01/2025).
[18]https://readcoop.eu/model/italian-general-model (accessed 2 9 /01/2025).
[19] Other available and open-source datasets can be found in the HTR-United platform, an aggregator of HTR datasets https://htr-united.github.io/catalog.html(accessed 29/01/2025) and as part of the project AI4Culture https://ai4culture.eu/resources?page=0&aiCategories=IMAGE_TO_TEXT&type=undefined&resourceType=all (accessed 29/01/2025).

| | | Century | | | dataset (accessed 14/01/2025) | |
|---|---|---|---|---|---|---|
| **ICDAR2017** | German, French, Italian | 19th Century | 500 (Train B, Batch 1) | Last 500 images of training ser | ICDAR2017 dataset (accessed 14/01/2025) | (Sánchez *et al.*, 2017) |

Table I: Basic information about the chosen Handwritten Text Recognition Datasets

### 3.3.1 An investigation of LLM pre-training using HTR datasets

The datasets mentioned are benchmarks that have been repeatedly used to assess and compare the performance of different models. As mentioned, Humphries et al. argued that HTR benchmark datasets were not suitable for pursuing the task in an unbiased manner, as they may have been used during the pre-training of LLMs due to their availability on the web (Humphries *et al.*, 2024, pp.12-13). However, they did not provide comprehensive evidence to support this claim. To demonstrate if these datasets have been part of pre-training sets, a similar approach to the one presented by Chang *et al.*, (2023) was followed, using a "name cloze" prompt, meaning a masked prompt which can effectively reveal if the material was completely memorized by the model. To run a similar test, 1% or, in the cases of small datasets, at least 5 pages per dataset were considered, creating sub-datasets (see Table II.), where, for each page, three random words with at least 4 characters were masked using [MASK].

| **Dataset** | **Pages used for the evaluation** |
|---|---|
| IAM | 5 |
| RIMES | 18 |
| Leopardi | 5 |
| LAM | 12 |
| Bentham | 5 |
| READ2016 | 5 |
| ICDAR2017 | 50 |

Table II: Number of pages included in each dataset to assess their potential use for pre-training.

Then, each of the chosen LLMs, was asked to complete the missing words using the following system prompt and user prompt inspired by Chang *et al.*, (2023):

> "You're an AI assistant specialized in filling masked words in a sentence. When a [MASK] token appears in the text, replace it with the most probable word, which has at least four characters. Consider the context carefully."
>
> "Text: {masked text} You've seen the passage in your training data. Replace each [MASK] with the most appropriate word, reply with ONLY ONE word even if you're not totally certain. Provide ONLY the replacement words, separated by spaces, in order of appearance."

Considering a total of 100 pages analyzed across datasets and three words masked per each dataset, the total number of predicted words was set to 300.

| **Model** | **Correct Words / Total Masked Words** |
|---|---|
| Gpt-4o-2024-08-06 | 17/300 |
| Claude-3-5-sonnet-20240620 | 27/300 |
| Qwen2-VL-7B | 2/300 |
| MinicpmV-2 6 | 1/300 |
| Pixtral-12B | 1/300 |
| InternVL2-8B | 1/300 |
| Phi-3-mini-128k-instruct | 0/300 |

Table III: Report of the correct masked predictions over the number of total masked words across datasets.

The results reported in Table III. show that the percentage of correct guesses over the masked words are minimal, which could suggest that they have not been used for pre-training LLMs. The case with the highest guesses is Claude Sonnet 3.5 which correctly guessed 15 out of 150 words in the ICDAR dataset, 6 out of 54 words in the RIMES, 2 out of 36 words in LAM, 2 out of 15 words in Bentham and 2 out of 15 words in READ.

## 3.4 *Experimental setup*

The implementation approach differs between proprietary to open-source models: the first can be leveraged via API, while open-source models must be locally installed using libraries such as HuggingFace[20]'s Transformers[21] or VLLM[22]. For this project, an NVIDIA L40S GPU with 48GB of GPU memory was used, which allowed it to run models up to 10–12 billion parameters. When possible, the local installation was performed using Flash Attention[23] (Dao *et al.*, 2022). Furthermore, the homogeneous adaptation of LLMs for the HTR task can be achieved by tuning the hyperparameters at the same value and maintaining the same prompt throughout the models. For this reason, the Temperature parameter, which adjusts the degree of randomness in the response, is set to zero (Humphries *et al.*, 2024, p.11). Whether the implementation via API supports directly the specification of Temperature, in the local installation the same results are achieved by setting the parameter "do sample" to False[24]. In this case, the model selects the token with the highest probability at each-step, following a greedy decoding approach. In the local implementations, the hyperparameter called repetition penalty[25] is often added to address issues of redundancy and looping repetitions in the generated text. In this case, it was set to 1.2, as it has been demonstrated to achieve a "good balance between truthful generation and lack of repetition" when combined with greedy sampling (Keskar *et al.*, 2019).

### 3.4.1 Zero-shot

Approaching prompting in a benchmark task means trying to find the general yet effective prompt that correctly suits the demands and the output's expected accuracy. To achieve this, the interaction with the model requires system and user prompt. The system prompt guides the model's responses by providing general instructions. By assigning the model a persona, following the "persona pattern" (White *et al.*, 2023, p.7), other than context-wise descriptions and steps to simplify the task, it learns to provide responses in a consistent manner. Moreover, the end of the system prompt generally is dedicated to specifying the customized output format[26]. For the benchmarking project, the chosen system prompt was:

> "You are an AI assistant specialized in transcribing handwritten text from images. Please follow these guidelines:
>
> 1. Examine the image carefully and identify all handwritten text.
> 2. Transcribe ONLY the handwritten text. Ignore any printed or machine-generated text in the image.
> 3. Maintain the original structure of the handwritten text, including line breaks and paragraphs.
> 4. Do not attempt to correct spelling or grammar in the handwritten text. Transcribe it exactly as written.
> 5. Do not describe the image or its contents.
> 6. Do not introduce or contextualize the transcription.
>
> Remember, your goal is to provide an accurate transcription of ONLY the handwritten portions of the text, preserving its original form as much as possible."

This prompt has been created by assigning the role of specialized transcription assistant, following a clear structure and separating distinctly the steps to be considered by the model, other than using a direct tone. In this case, the repetition of certain instructions is fundamental, particularly when added at the end of the prompt, to specify what the model needs to pay specific attention to[27]. On the other hand, the user prompt represents the direct user-agent interaction and in this case, it was formulated as follows:

> "Please transcribe the handwritten text in this image as accurately as possible, respecting line breaks. Every response should start with "Transcription:", followed only by the transcription."

This baseline prompt defines the task in a straightforward manner and has been proven effective and suitable for transcriptions across various datasets. The consistency of the output is determined by the guide added at the end of the input, which specifies how to initiate the output. This approach, combined with the guidelines

---

[20] https://huggingface.co/ (accessed 29/01/2025)
[21] https://huggingface.co/docs/transformers/index (accessed 29/01/2025)
[22] https://docs.vllm.ai/en/latest/ (accessed 29/01/2025)
[23] https://huggingface.co/docs/text-generation-inference/conceptual/flash_attention (accessed 29/01/2025)
[24] https://huggingface.co/docs/transformers/generation_strategies (accessed 29/01/2025)
[25] https://huggingface.co/docs/transformers/main_classes/text_generation#transformers.GenerationConfig.repetition_penalty (accessed 29/01/2025)
[26] https://promptengineering.org/system-prompts-in-large-language-models/ (accessed 29/01/2025).
[27] https://huggingface.co/docs/transformers/main/tasks/prompting (accessed 29/01/2025)

provided in the system prompt, definitively eliminates polite intro- ductions typically generated by the model at the beginning of the predicted text. Consequently, by assuming that outputs started always with the same pattern, the expression was then manually deleted. Specific output formatting is demonstrated to enhance significantly the performance of the models, for instance, the JSON format is recommended for increasing consistency[28]. In this case, however, the output format chosen was plaintext, because, unlike OpenAI models, not all the selected models have a JSON mode integrated and simply specifying the desired output format in the prompt, even with a pre-structured template, was not guaranteeing a uniform output.

### 3.4.2 LLMs post-correction

Post-OCR or HTR correction can be approached in different ways, from crowdsourcing to the automatic post-correction of previous predictions using LLMs (Bourne, 2025, p.2). In this context, LLMs are tested for their ability to perform intrinsic self-correction, where the model autonomously adjusts its responses without external human feedback (Huang *et al.*, 2024, p.1), based on the assumption that verifying correctness is easier than generating text from scratch.
The aim of this part of the project is to determine whether an increase in accuracy could be observed in the predicted text after LLM post-correction. The refinement task is carried out in three steps where the model is asked again to analyze both the original image and the output produced at time t – 1, which means that in the first iteration the model examines the output of the zero-shot task. In each subsequent step a different user prompt is provided, specifying where the focus of the model is needed, whether in the orthography or spelling refinement or in the layout formatting or in both as the last prompt:

> "Review the original image and your previous transcription. Focus on correcting any spelling errors, punctuation mistakes, or missed words. Ensure the transcription accurately reflects the handwritten text. Every response should start with 'Transcription:', followed only by the transcription.",
>
> "Examine the structure of the transcription. Are paragraphs and line breaks correctly represented? Adjust the layout to match the original handwritten text more closely. Every response should start with 'Transcription:', followed only by the transcription.",
>
> "Make a final pass over the transcription, comparing it closely with the original image. Make any last corrections or improvements to ensure the highest possible accuracy. Every response should start with 'Transcription:', followed only by the transcription."

## *3.5 Evaluation*

For an evaluation of the results, classic metrics borrowed from the Automatic Speech Recognition field such as Character Error Rate, Word Error Rate, and Word Error Rate-Bag of Words are adopted, as they have been widely used in literature, previous benchmark assessments, and in *Transkribus* to evaluate the accuracy of the predictions. These metrics provide a comprehensive evaluation by considering both the reading order, in the case of CER and WER, or not as in the case of WER-Bag of Words approach. This combination allows for the assessment of the transcription quality and the model's ability to provide a prediction in which meaningful information can be extracted, even when an acceptable level of transcription is not reached. Before computing these metrics, the output length is normalized to the one of the GT, while line breaks, empty lines and all punctuation followed by a whitespace are removed. This ensures that whitespaces are the only separators between words, allowing long text blocks to be formatted correctly for evaluation, focusing purely on the content rather than any formatting discrepancies.

- Character Error Rate and Word Error Rate

  Character Error Rate is the "inverted accuracy", meaning that it represents the error rate at character level based on the Levenshtein distance (Levenshtein, 1965) This suggests that all the metrics that incorporate this aspect depend on the reading order. To calculate the CER, the following formula has been used:

$$CER = \frac{I + S + D}{N} = \frac{I + S + D}{C + S + D}$$

Where N is the total number of characters in the GT, C is the number of correct characters, I is the number of insertions, S of substitutions and D deletions required to transform the reference text into the provided GT (Neudecker *et al.*, 2021, p.2). To give an idea of what these percentages actually represent, a CER below 5% is considered very good, if it falls in a range between 5 to 10% is good (Muehlberger *et al.*, 2019, p.965), excellence is achieved with a CER below 2.5% (Hodel *et al.*, 2021,

[28] https://platform.openai.com/docs/guides/structured-outputs (accessed 29/01/2025).

p.2).

On the contrary, the score below 90% of accuracy, meaning a CER above 10% it's an indication of poor quality (Ströbel *et al.*, 2022, p.1). Instead, Word Error Rate is the correspondent to CER but at word level, identifying the number of insertions, deletions and substitutions in the recognized words divided by the total number of words in GT. In this experiment, the WER and the derived WER-BoW computation is case insensitive, meaning that uppercase letters and lowercase letters are considered equal.

- Bag of Words - Word Error Rate

The previously introduced metrics are based on the reading order, meaning that, if a word is detected but not in the correct positioning, it will not be considered. To address these issues and consider the correctly detected words regardless of their order, which is particularly useful for information retrieval tasks, the WER-Bag of Words (WER-BoW) metric was introduced (Moysset *et al.*, 2017; Pletschacher *et al.*, 2015). In this case, WER-BoW is calculated considering the difference between the total number of words in the GT and the intersection of those present in both the GT and in the prediction, over the total number of words:

$$WER - BoW = \frac{N - (P \cap N)}{N}$$

Where N is the total number of words in the GT and P the number of words in the prediction. Therefore, this value can be much lower than WER when the reading order between GT and prediction is different.

# 4 Analysis of Results and Discussion

Tables IV to X present the zero-shot results, indicating that LLMs perform well in transcribing English handwritten text and modern handwriting, though their accuracy declines progressively in other languages. While proprietary models, particularly Claude Sonnet 3.5, continue to yield the best results, open-source models like Qwen are narrowing the gap. However, it cannot be broadly stated that *Transkribus* models consistently outperform LLMs, as accuracy varies depending on the specific use case.
A clear disparity emerges between the recognition of modern and historical handwriting. In modern handwriting, LLMs surpass Transkribus, achieving excellent results with a CER below 5%. On IAM, GPT-4o-mini attains 1.71% CER and 3.34% WER, while on RIMES, GPT-4o reaches 1.69% CER and 3.66% WER, outperforming T*ranskribus*' supermodel.
Instead, for the recognition of the English historical dataset, the results are balanced. While The Text Titan achieves the best outcome with 7,07% CER and 12,41% WER, LLMs are not far behind. However, the Bentham dataset was used in the pre-training of existing *Transkribus PyLaia* models, consequently, it can be supposed that even supermodels could have been trained using this data(Muehlberger *et al.*, 2019, p.959). Moreover, the discrepancy between the models' performance dealing with English historical and modern texts exhibits a language bias which mirrors the intrinsic one in LLMs caused by most of its training data being in English. Therefore, these models result in being biased not only from a linguistic aspect but also in relation to the characteristics of handwriting (Hodel, 2022, p.169). The accuracy decline is even more pronounced for non-English datasets, where neither *Transkribus* models nor LLMs consistently outperform one another. For Italian, Claude Sonnet 3.5 delivers the best results on both the Leopardi and LAM datasets, yet the overall performance remains poor, with CER exceeding 20%. On LAM, Claude reaches 20.55% CER and 27.78% WER, while on Leopardi, it records 26.43% CER and 35.08% WER. In German and multilingual datasets, The Text Titan surpasses LLMs, achieving 40.63% CER and 64.28% WER on the READ2016 dataset and 14.40% CER and 29.40% WER on ICDAR2017. *Transkribus* also struggles to produce satisfactory results, likely due to the challenges of difficult handwriting, conservation issues, and the complexity of historical German lexicon.

| Model | CER | WER | WER BoW |
|---|---|---|---|
| The Text Titan I | 9.13% | 23.38% | 17.88% |
| Gpt-4o-mini-2024-07-18 | **1.71%** | **3.34%** | **2.33%** |
| Gpt-4o-2024-08-06 | 1.75% | 3.59% | 2.49% |
| Claude-3-5-sonnet-20240620 | 1.75% | 3.55% | 2.57% |
| MinicpmV-2 6 | 2.02% | 3.23% | 2.42% |
| Qwen2-VL-7B | 2.30% | 4.20% | 2.43% |
| Pixtral-12B | 2.92% | 5.39% | 3.54% |
| InternVL2-8B | 25.15% | 39.95% | 33.05% |
| Phi-3-mini-128k-instruct | 3.85% | 6.40% | 4.46% |

Table IV: IAM dataset results.

| Model | CER | WER | WER BoW |
|---|---|---|---|
| The Text Titan I | 10.71% | 20.76% | 17.93% |
| Gpt-4o-mini-2024-07-18 | 3.22% | 7.68% | 6.29% |
| Gpt-4o-2024-08-06 | 1.69% | **3.76%** | **3.30%** |
| Claude-3-5-sonnet-20240620 | **1.63%** | 4% | 3.36% |
| MinicpmV-2 6 | 7.98% | 16.55% | 15% |
| Qwen2-VL-7B | 6.47% | 12.62% | 10.42% |
| Pixtral-12B | 13.02% | 23.55% | 18.73% |
| InternVL2-8B | 35.19% | 66.53% | 57.61% |
| Phi-3-mini-128k-instruct | 23.58% | 48.45% | 43.60% |

Table V: RIMES dataset results.

| Model | CER | WER | WER BoW |
|---|---|---|---|
| Italian Handwriting M1 | 25.94% | 45.77% | 36.75% |
| Gpt-4o-mini-2024-07-18 | 37.86% | 60.83% | 52.25% |
| Gpt-4o-2024-08-06 | 33.53% | 49.81% | 42.53% |
| Claude-3-5-sonnet-20240620 | **20.55%** | **27.78%** | **24.57%** |
| MinicpmV-2 6 | 41.55% | 62.56% | 61.05% |
| Qwen2-VL-7B | 28.76% | 42.46% | 40.80% |
| Pixtral-12B | 52.73% | 64.40% | 62.44% |
| InternVL2-8B | 80.82% | 94.22% | 93.24% |
| Phi-3-mini-128k-instruct | 64.78% | 81.42% | 80.12% |

Table VI: LAM dataset results.

| Model | CER | WER | WER BoW |
|---|---|---|---|
| Italian Handwriting M1 | 37.23% | 55.77% | 47.34% |
| Gpt-4o-mini-2024-07-18 | 48.13% | 67.82% | 55.46% |
| Gpt-4o-2024-08-06 | 36.34% | 51.67% | 43.82% |
| Claude-3-5-sonnet-20240620 | **26.43%** | **35.08%** | **31.15%** |
| MinicpmV-2 6 | 45.85% | 65.73% | 60.82% |
| Qwen2-VL-7B | 37.07% | 49.07% | 43.91% |
| Pixtral-12B | 59.54% | 82.02% | 68.50% |
| InternVL2-8B | 83.43% | 99.70% | 97.11% |
| Phi-3-mini-128k-instruct | 69.78% | 85.06% | 82.01% |

Table VII: Leopardi dataset results.

| Model | CER | WER | WER BoW |
|---|---|---|---|
| The Text Titan I | **7.07%** | **12.41%** | **8.54%** |
| Gpt-4o-mini-2024-07-18 | 9.48% | 15.09% | 13.16% |
| Gpt-4o-2024-08-06 | 16.62% | 20.73% | 18.89% |
| Claude-3-5-sonnet-20240620 | 10.97% | 14.46% | 12.24% |
| MinicpmV-2 6 | 11.76% | 17.24% | 13.91% |
| Qwen2-VL-7B | 8.01% | 12.94% | 11% |
| Pixtral-12B | 28.08% | 38.25% | 30.32% |
| InternVL2-8B | 76.81% | 95.92% | 81.67% |
| Phi-3-mini-128k-instruct | 32.03% | 41.85% | 38.73% |

Table VIII: Bentham dataset results.

| Model | CER | WER | WER BoW |
|---|---|---|---|
| The Text Titan I | **40.63%** | **64.28%** | **61.29%** |
| Gpt-4o-mini-2024-07-18 | 78.51% | 98.27% | 95.77% |
| Gpt-4o-2024-08-06 | 80.20% | 98.08% | 95.46% |
| Claude-3-5-sonnet-20240620 | 71.17% | 95.39% | 92.01% |
| MinicpmV-2 6 | 81.33% | 99.39% | 98.84% |
| Qwen2-VL-7B | 76.37% | 97.91% | 96.84% |
| Pixtral-12B | 77.88% | 99.48% | 98.20% |
| InternVL2-8B | 81.62% | 99.76% | 99.08% |
| Phi-3-mini-128k-instruct | 92.32% | 99.95% | 99.71% |

Table IX: READ dataset results.

| Model | CER | WER | WER BoW |
|---|---|---|---|
| The Text Titan I | **14.40%** | **29.47%** | **24.69%** |
| Gpt-4o-mini-2024-07-18 | 58.94% | 82.08% | 70.68% |
| Gpt-4o-2024-08-06 | 60.98% | 76.64% | 70.35% |
| Claude-3-5-sonnet-20240620 | 41.19% | 60% | 51.69% |
| MinicpmV-2 6 | 71.16% | 92.95% | 89.95% |
| Qwen2-VL-7B | 58% | 81.55% | 72.53% |
| Pixtral-12B | 71.88% | 95.40% | 82.05% |
| InternVL2-8B | 81.82% | 99.65% | 97.80% |
| Phi-3-mini-128k-instruct | 85.46% | 97.40% | 95.88% |

Table X: ICDAR2017 dataset results.

The results derived from the LLMs' post-correction, reported from Table XI to XVII, demonstrate how these models cannot guarantee a substantial improvement of the first prediction. While small improvements in performance are observed in some cases, this behavior is not consistent. This means that the same model applied to different datasets does not always show the same performance: it may correct errors when applied to some datasets but worsen performance in others. When improvements do occur, they are generally insignificant and insufficient to shift the output from "unusable" (CER over 10%) to "good". In fact, the best case of post-correction improvement was seen in GPT-4o applied to ICDAR2017 dataset noticing a decrease of 8% in CER and 4,7% in WER, but due to the high levels of error, these transcriptions remain unusable. Moreover, the models which demonstrated some capabilities of improvement are GPT models and Claude Sonnet 3.5 as also noted by Bourne (Bourne, 2025, p.14). Instead, all the open-source models failed to show improvement and rather increased the error rates. From this second sub-task inside the benchmarking project, it can be concluded that, aligning with Humphries *et al.*(2024), LLMs post correction does not lead to substantial prediction improvements and cannot be considered as a valid substitute for manual post correction at this moment.

| Model | CER | WER | WER BoW |
|---|---|---|---|
| Gpt-4o-mini-2024-07-18 | 1.74% | **2.86%** | **1.86%** |
| Gpt-4o-2024-08-06 | **1.39%** | 3.46% | 2.48% |
| Claude-3-5-sonnet-20240620 | 8.55% | 10.28% | 5.62% |
| MinicpmV-2 6 | 2.02% | 3.27% | 2.34% |
| Qwen2-VL-7B | 5.06% | 9% | 3.60% |
| Pixtral-12B | 10.03% | 12.04% | 6.94% |
| InternVL2-8B | 24.74% | 39.11% | 32.82% |
| Phi-3-mini-128k-instruct | 3.66% | 7.09% | 5.08% |

Table XI: IAM dataset results.

| Model | CER | WER | WER BoW |
|---|---|---|---|
| Gpt-4o-mini-2024-07-18 | 3.45% | 8.02% | 6.28% |
| Gpt-4o-2024-08-06 | 1.74% | **4.17%** | 3.69% |
| Claude-3-5-sonnet-20240620 | **1.61%** | **4.17%** | **3.51%** |
| MinicpmV-2 6 | 15.51% | 19.59% | 16.32% |
| Qwen2-VL-7B | 10.23% | 20.33% | 15.44% |
| Pixtral-12B | 16.92% | 34.25% | 29.15% |
| InternVL2-8B | 35.05% | 66.24% | 57.94% |
| Phi-3-mini-128k-instruct | 33.23% | 59.57% | 55.31% |

Table XII: RIMES dataset results.

| Model | CER | WER | WER BoW |
|---|---|---|---|
| Gpt-4o-mini-2024-07-18 | 37.92% | 52.98% | 44.77% |
| Gpt-4o-2024-08-06 | 36.56% | 52.32% | 46.71% |
| Claude-3-5-sonnet-20240620 | **20.08%** | **34.43%** | **24.56%** |
| MinicpmV-2 6 | 43.62% | 68.84% | 61.59% |
| Qwen2-VL-7B | 31.11% | 52.11% | 44.25% |
| Pixtral-12B | 54.17% | 79.25% | 64.69% |
| InternVL2-8B | 80.04% | 98.41% | 92.81% |
| Phi-3-mini-128k-instruct | 73.77% | 93.18% | 90.09% |

Table XIII: LAM dataset results.

| Model | CER | WER | WER BoW |
|---|---|---|---|
| Gpt-4o-mini-2024-07-18 | 47.36% | 67.99% | 54.91% |
| Gpt-4o-2024-08-06 | 36.25% | 52.46% | 44.32% |
| Claude-3-5-sonnet-20240620 | **26%** | **34.20%** | **30.52%** |
| MinicpmV-2 6 | 47.04% | 65.87% | 60.18% |
| Qwen2-VL-7B | 40.08% | 54.37% | 46.96% |
| Pixtral-12B | 56.89% | 81.23% | 68.82% |
| InternVL2-8B | 83.27% | 99.67% | 98.89% |
| Phi-3-mini-128k-instruct | 76.86% | 91.95% | 88.28% |

Table XIV: Leopardi dataset results.

| Model | CER | WER | WER BoW |
|---|---|---|---|
| Gpt-4o-mini-2024-07-18 | 60.57% | 82.50% | 70.60% |
| Gpt-4o-2024-08-06 | 52.23% | 72.16% | 63.41% |
| Claude-3-5-sonnet-20240620 | **40.87%** | **59.41%** | **50.96%** |
| MinicpmV-2 6 | 69.35% | 92.47% | 89.29% |
| Qwen2-VL-7B | 60.57% | 83.71% | 73.56% |
| Pixtral-12B | 72.32% | 95.36% | 82.31% |
| InternVL2-8B | 80.78% | 99.70% | 97.58% |
| Phi-3-mini-128k-instruct | 77.98% | 97.63% | 95.54% |

Table XV: Bentham dataset results.

| Model | CER | WER | WER BoW |
|---|---|---|---|
| Gpt-4o-mini-2024-07-18 | 78.22% | 97.23% | 95.69% |
| Gpt-4o-2024-08-06 | 77.42% | 97.59% | 94.19% |
| Claude-3-5-sonnet-20240620 | **71.08%** | **95.53%** | **92.17%** |
| MinicpmV-2 6 | 80.74% | 99.74% | 99.24% |
| Qwen2-VL-7B | 76.25% | 97.77% | 96.78% |
| Pixtral-12B | 79% | 99% | 98.03% |
| InternVL2-8B | 82.34% | 99.89% | 99.03% |
| Phi-3-mini-128k-instruct | 82.31% | 99.77% | 99.42% |

Table XVI: READ2016 dataset results.

| Model | CER | WER | WER BoW |
|---|---|---|---|
| Gpt-4o-mini-2024-07-18 | 78.22% | 97.23% | 95.69% |
| Gpt-4o-2024-08-06 | 77.42% | 97.59% | 94.19% |
| Claude-3-5-sonnet-20240620 | **71.08%** | **95.53%** | **92.17%** |
| MinicpmV-2 6 | 80.74% | 99.74% | 99.24% |
| Qwen2-VL-7B | 76.25% | 97.77% | 96.78% |
| Pixtral-12B | 79% | 99% | 98.03% |
| InternVL2-8B | 82.34% | 99.89% | 99.03% |
| Phi-3-mini-128k-instruct | 82.31% | 99.77% | 99.42% |

Table XVII: ICDAR2017 dataset results.

# 5 Conclusion and Future Developments

In conclusion, this paper has analyzed the task of Handwritten Text Recognition, proposing an initial exploration of the applicability of MLLMs in a multilingual HTR context using a snapshot of models available as of October 2024. To assess this potential, the research conducted a benchmarking study that examined LLMs' zero-shot capabilities from a user's perspective, including hyperparameter choices and evaluation metrics. Additionally, LLMs were prompted to self-correct their primary output, evaluating whether post-correction could be feasible without manual intervention.

It emerged that LLMs applied to HTR offer several advantages, including ease of implementation, improved user-model interaction, faster processing times, and reduced costs. The differences in workflow, when compared to traditional approaches, could significantly alter how this task is adapted, potentially enabling a single general model to recognize various handwriting styles and languages. Such advancements could enhance HTR predictions and promote a wider adoption in digital libraries.

However, the results of this research show that the feasibility of using both proprietary and open- source LLMs for HTR is skewed towards the English language and mostly on modern handwriting documents, caused by the proportionally unbalanced datasets used during pre-training. Consequently, the performance on other languages and historical documents is consistently weaker, generally producing unusable results. The model which constantly demonstrated the best results overall is Claude Sonnet 3.5. While the accuracy is similar between proprietary and open-source models on modern handwriting and English materials, open-source model performance decreases significantly for historical documents in other languages. Moreover, MLLMs do not demonstrate a consistent and significant capability of autocorrection. In particular, it can be observed that post-corrections produced by open-source models reduced accuracy overall. As for the comparison with *Transkribus*' models, it is not possible to generalize if the platform's models outperform LLMs or vice versa. While LLMs achieved comparable results for English historical handwriting and outperformed *Transkribus* on modern handwriting and Italian datasets, *Transkribus* models showed better results on German and multilingual datasets.

Platforms like *Transkribus* and general LLMs will likely continue to coexist as tools supporting users' activities, each being selected based on specific needs. LLMs are quicker, less expensive in terms of material preparation and adaptation, and allow for iterative task adjustments through interaction with the API. However, they still require improvement in the recognition of historical handwritten documents in different languages. On the other hand, *Transkribus* offers a wide variety of tools, and the shift from highly specialized models to supermodels will likely lead to uniform improved performance on languages other than English. At the moment, for tasks requiring highly tailored solutions, *Transkribus*' user interface and specialized models remain advantageous.

This study was intended as an analysis of the state of these models at the time of publishing. Its limitations include the choice of datasets, which was limited to available HTR datasets in a few languages and does not fully capture the models' performance in multilingual contexts. The benchmarking approach provided a basis for comparison, but the baseline prompt lacked contextual elements that could have significantly improved the results. Furthermore, the analysis was limited to open-source models up to 12B parameters due to computational constraints.

Future improvements in both proprietary and open-source models will surely produce better results, especially given the rapid advancements in the field of Multimodal LLMs. To ensure broader representability and generalizable outcomes, future research should involve diverse HTR datasets, including ideographic and non-Latin scripts. Additionally, exploring alternative prompt structures, such as contextual prompting, few-shot learning, or incorporating fine-tuning, could further improve performance. Fine-tuning LLMs for HTR tasks could specialize these models to a com- parable level to *Transkribus*. A next relevant step would be to investigate whether a comparison between specialized LLMs and *Transkribus*' models would produce results consistent with the findings presented in this study.

*Acknowledgments*
We would like to thank *I Tatti, The Harvard University Center for Italian Renaissance Studies* for providing access to the servers, subscriptions, and GPUs necessary for the execution of the project.